\documentclass[11pt]{article}

\usepackage{url}
\usepackage{caption}
\usepackage{moresize}
\usepackage{lastpage}
\usepackage{amsmath}
\usepackage{graphicx}
\usepackage{multirow}
\usepackage{multicol}

\usepackage{algorithm}
\usepackage[noend]{algpseudocode}
\usepackage{setspace}
\let\Algorithm\algorithm
\renewcommand\algorithm[1][]{\Algorithm[#1]\setstretch{1.1}}

\algrenewcommand{\algorithmiccomment}[1]{\hskip3px$\#$ #1}
\algdef{SE}[SUBALG]{Indent}{EndIndent}{}{\algorithmicend\ }%
\algtext*{Indent}
\algtext*{EndIndent}
\algnewcommand{\LineComment}[1]{\hspace{-15pt}$\#$ #1}

\usepackage{hyperref}
\usepackage{xcolor}
\hypersetup{
    colorlinks=true,
    linkcolor=blue,
    filecolor=magenta,      
    urlcolor=olive,
    citecolor=cyan,
}

\newcommand{\sq}{\textquotesingle}
\newcommand\reps{96,192 }
\newcommand\runs{28,857,600 }

\title{High Per Parameter:\\A Large-Scale Study of Hyperparameter Tuning\\for Machine Learning Algorithms}

\author{Moshe Sipper}

\begin{document}
\maketitle

\begin{abstract}
Hyperparameters in machine learning (ML) have received a fair amount of attention, and hyperparameter tuning has come to be regarded as an important step in the ML pipeline.
But just how useful is said tuning? While smaller-scale experiments have been previously conducted, herein we carry out a large-scale investigation, specifically, one involving 26 ML algorithms, 250 datasets (regression and both binary and multinomial classification), 6 score metrics, and \runs algorithm runs. Analyzing the results we conclude that for many ML algorithms we should not expect considerable gains from hyperparameter tuning \textit{on average}, however, there may be some datasets for which default hyperparameters perform poorly, this latter being truer for some algorithms than others. By defining a single \textit{hp\_score} value, which combines an algorithm's accumulated statistics, we are able to rank the 26 ML algorithms from those expected to gain the most from hyperparameter tuning to those expected to gain the least. We believe such a study may serve ML practitioners at large. 
\end{abstract}

% Keywords: Machine Learning, Hyperparemeters 

\section{Introduction}
\label{sec:intro}

In machine learning (ML), a hyperparameter is a parameter whose value is given by the user and used to control the learning process. This is in contrast to other parameters, whose values are obtained algorithmically via training.

While software packages invariably provide hyperparameter defaults, practitioners will often tune these---either manually or through some automated process---to gain better performance. 

Herein, we propose to examine the issue of hyperparameter tuning through an extensive empirical study, involving multitudinous algorithms, datasets, metrics, and hyperparameters. Our aim is to assess just how much of a performance gain can be had per algorithm by employing a performant tuning method.

The next section presents a brief account of previous work. Section~\ref{sec:exp} describes the experimental setup, followed by results and discussion in Section~\ref{sec:results}.
Finally, we end with concluding remarks in Section~\ref{sec:disc}.

\section{Previous Work}
\label{sec:prev}

There has been a fair amount of work on hyperparameters and it is beyond this paper's scope to provide a review. For that, we refer the reader to the recent, comprehensive review: ``Hyperparameter Optimization: Foundations, Algorithms, Best Practices and Open Challenges'' \cite{Bischl2021}. They wrote that, ``we would like to tune as few HPs [hyperparameters] as possible. If no prior knowledge from earlier experiments or expert knowledge exists, it is common practice to leave other HPs at their software default values...''

\cite{Bischl2021} also noted that, ``more sophisticated HPO [hyperparameter optimization] approaches in particular are not as widely used as they could (or should) be in practice.'' We shall use a sophisticated HPO approach herein. The paper does not include an empirical study.

We present below only the most recent papers that are directly relevant to the current study.

A major work by \cite{Probst2019} formalized the problem of hyperparameter tuning from a statistical point of view, defined data-based defaults, and suggested general measures quantifying the tunability of hyperparameters. The overall tunability of an ML algorithm or that of a specific hyperparameter was essentially defined by comparing the gain attained through tuning with some baseline performance, usually that attained when using default hyperparameters. They also conducted an empirical study involving 38 binary classification datasets from OpenML, and six ML algorithms: elastic net, decision tree, k-nearest neighbors, support vector machine, random forest, and xgboost. Tuning was done through random search.
They found that some algorithms benefited from tuning more than others, with elastic net and svm showing the most improvement and random forest showing the least.

\cite{Hilde2020} presented a methodology to determine the importance of tuning a hyperparameter based on a non-inferiority test and tuning risk: the performance loss that is incurred when a hyperparameter is not tuned, but set to a default value. They performed an empirical study involving 59 datasets from OpenML and two ML algorithms: support vector machine and random forest. Tuning was done through random search. Their results showed that leaving particular hyperparameters at their default value is noninferior to tuning these hyperparameters. In some cases, leaving the hyperparameter at its default value even outperformed tuning it. 

Finally, \cite{Turner2020} recently presented results and insights pertaining to the black-box optimization (BBO) challenge at NeurIPS 2020. Analyzing the performance of 65 submitted entries, they concluded that, ``Bayesian optimization is superior to random search for machine learning hyperparameter tuning'' (indeed this is the paper's title). (NB: random search is usually better than grid search, e.g., \cite{Bergstra2012}). We shall use Bayesian optimization herein.

Examining these recent studies we made the following decisions regarding our setup, which is described in the next section:
\begin{itemize}
    \item Consider significantly more algorithms.
    \item Consider significantly more datasets.
    \item Consider Bayesian optimization, rather than lesser-performing random search or grid search.
\end{itemize}

\section{Experimental Setup}
\label{sec:exp}

Our setup involves numerous runs across a plethora of algorithms and datasets, comparing tuned and untuned performance over six distinct metrics. Below we detail the following setup components: datasets,  algorithms, metrics, hyperparameter tuning, overall flow.

\begin{figure}
    \centering
    \begin{tabular}{cc}
     \includegraphics[width=0.47\textwidth]{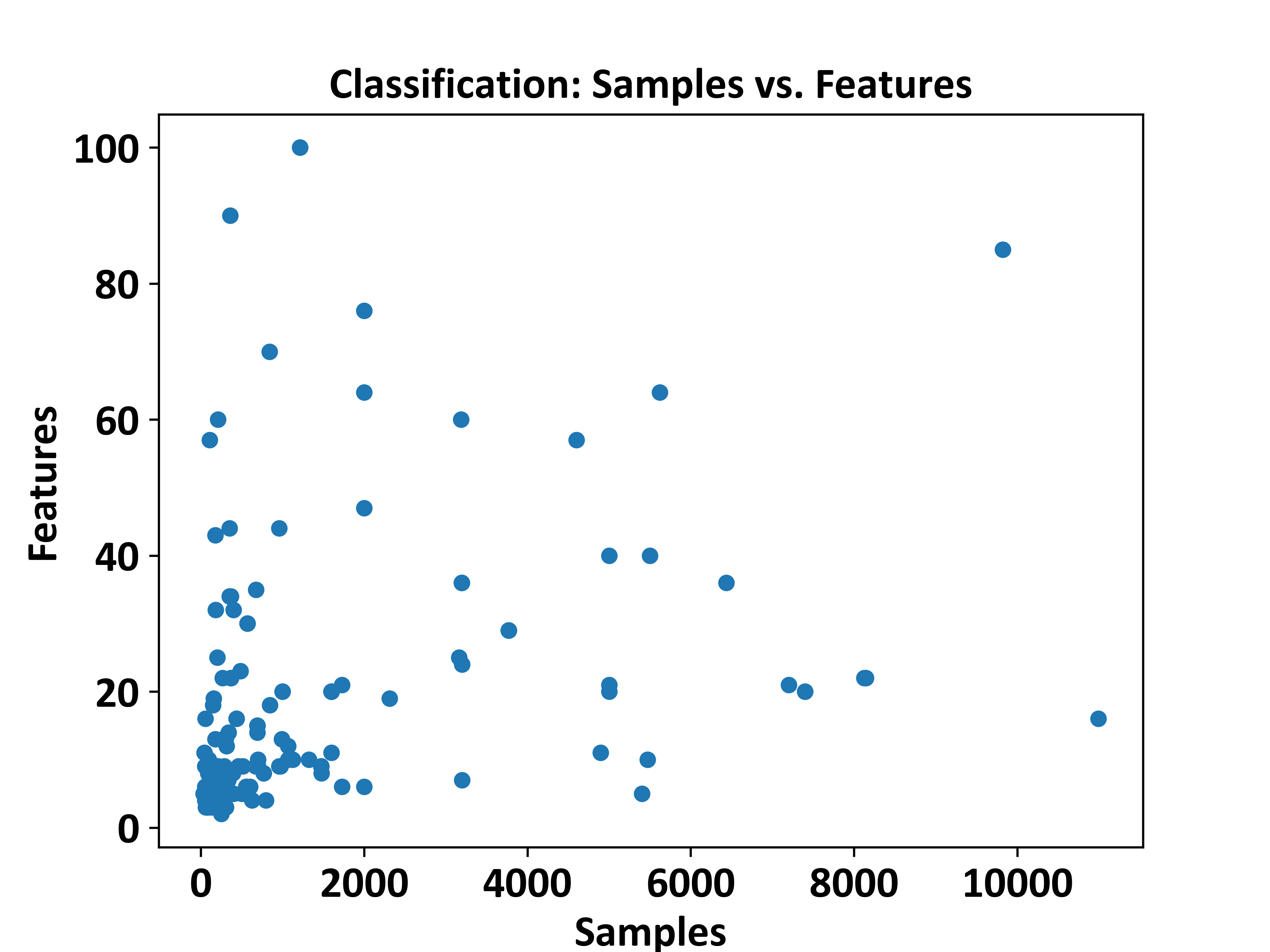} & \includegraphics[width=0.47\textwidth]{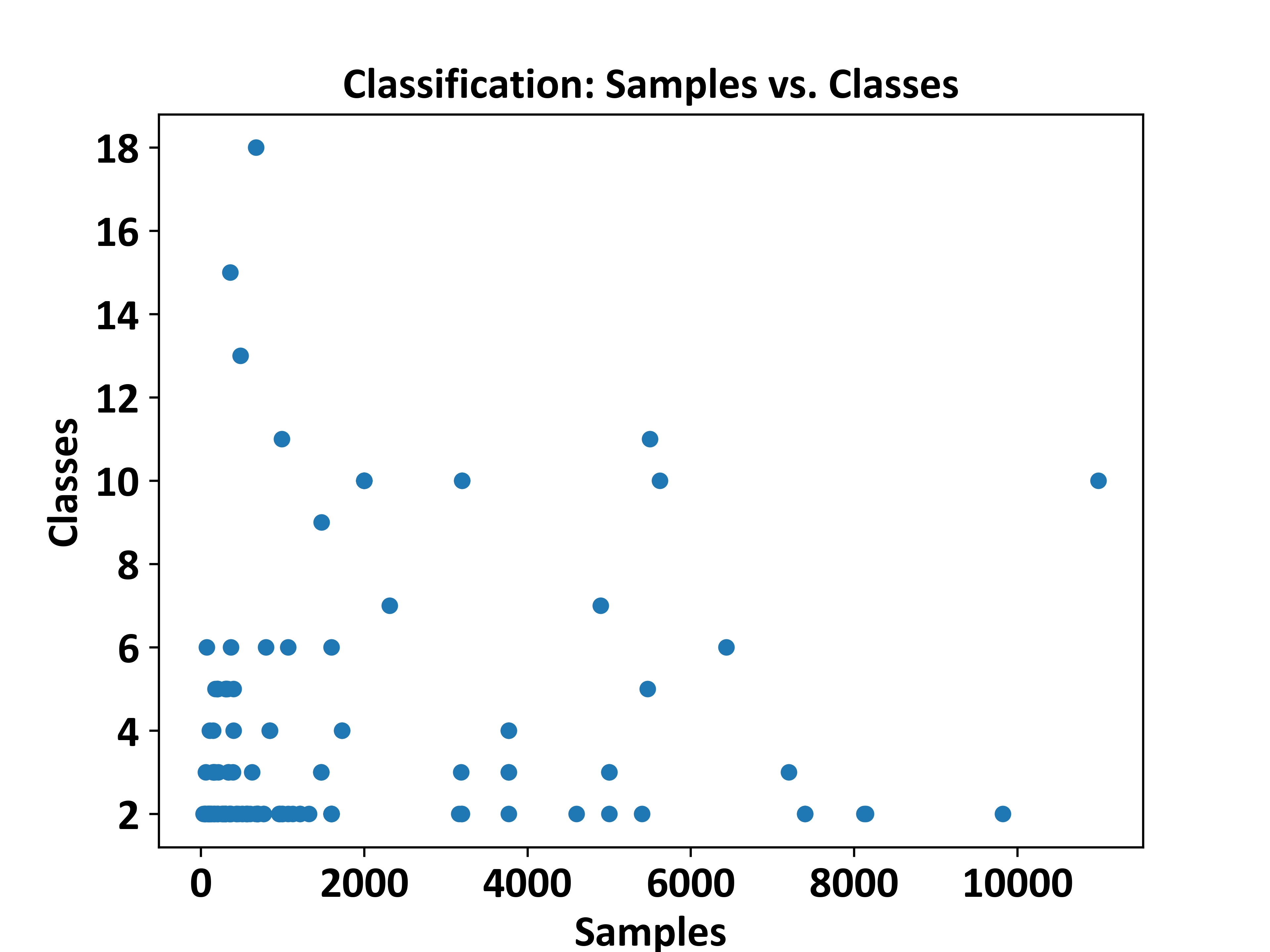} \\
     \includegraphics[width=0.47\textwidth]{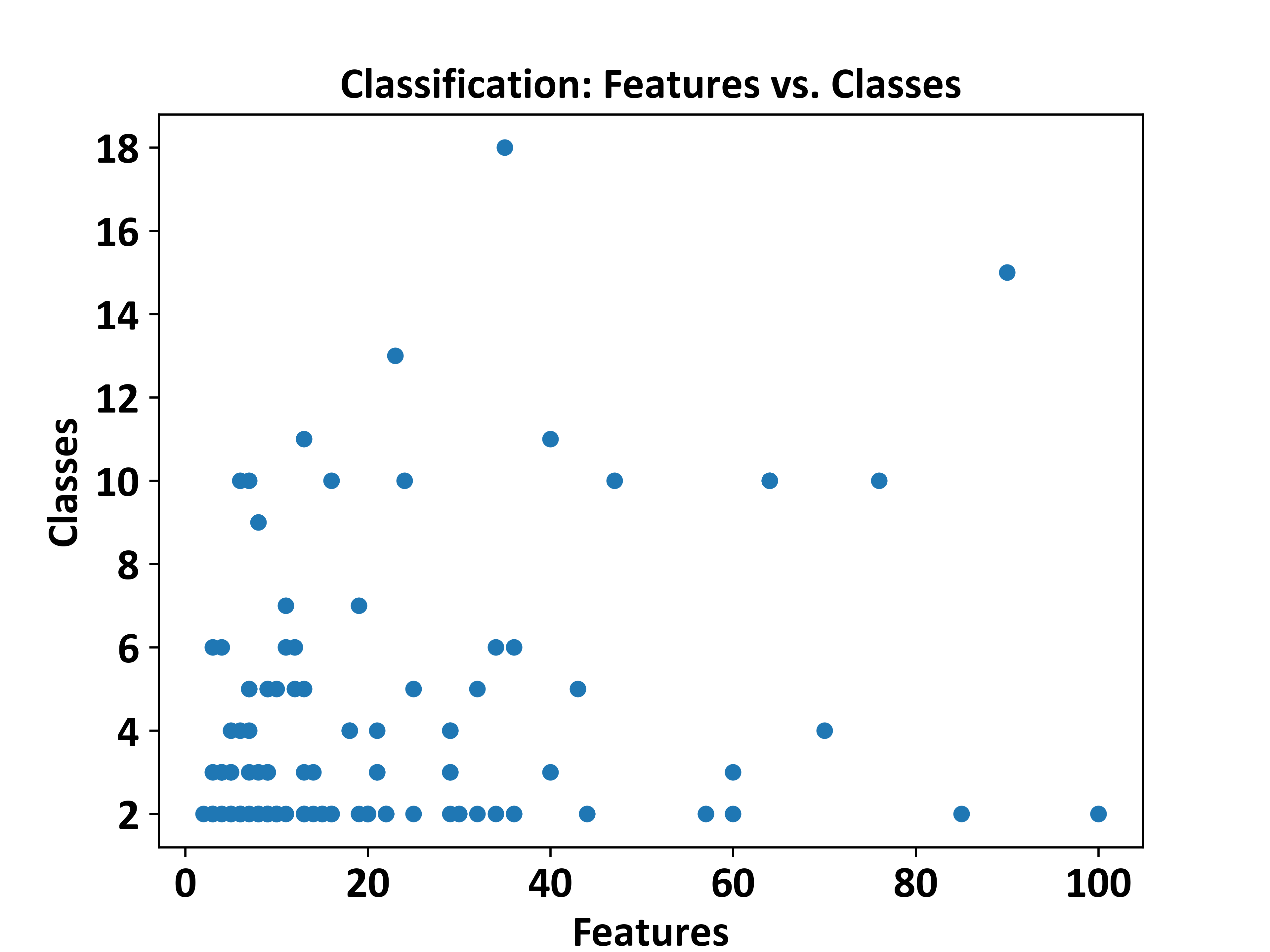} & \includegraphics[width=0.47\textwidth]{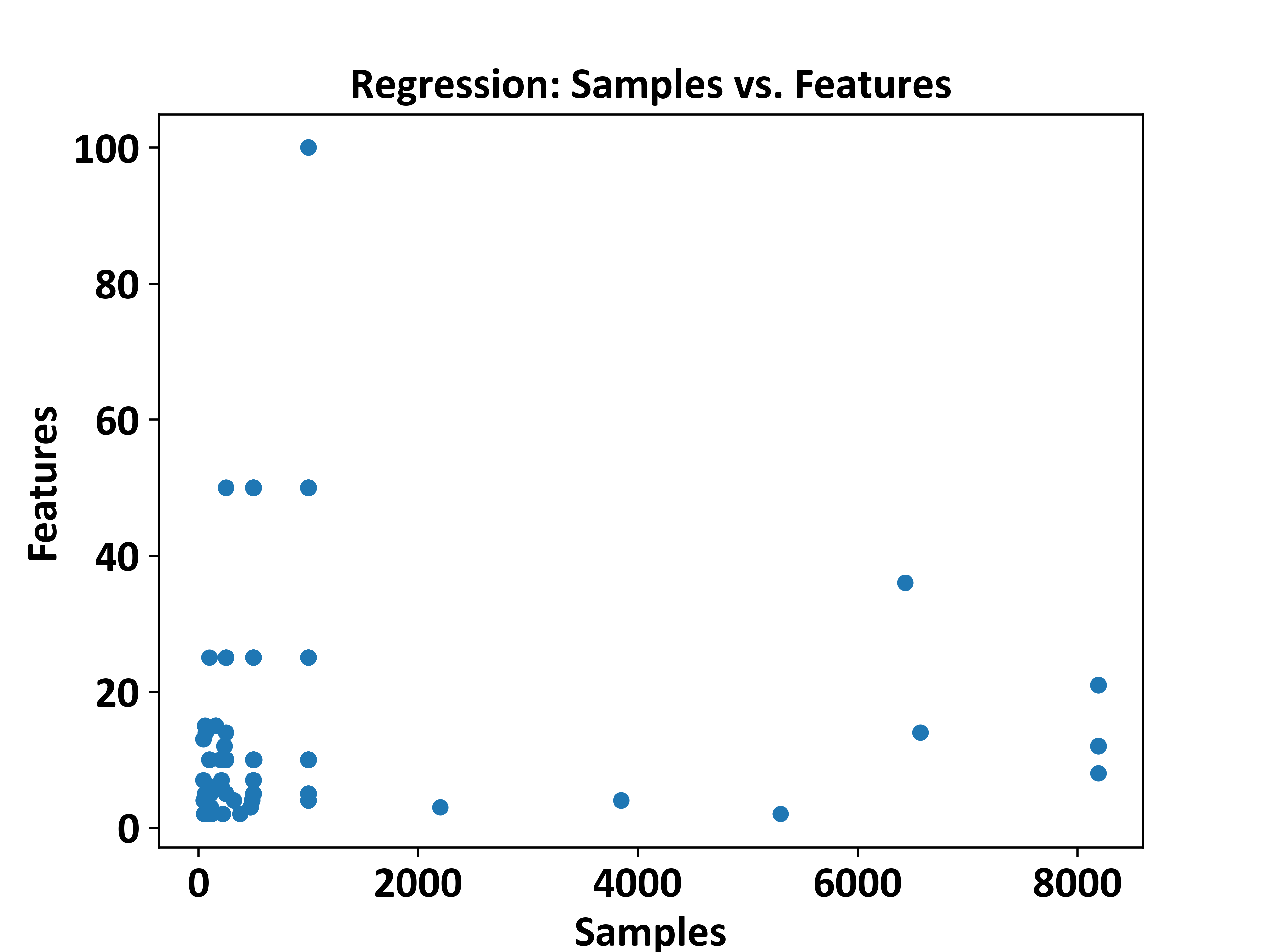}  \\
    \end{tabular}
    \caption{Characteristics of the 144 classification datasets and 106 regression datasets used in our study.}
    \label{fig:ds}
\end{figure}

\paragraph{Datasets.}
We used the recently introduced PMLB repository \cite{romano2021pmlb}, which includes 166 classification datasets and 122 regression datasets. As we were interested in performing numerous runs, we retained the 144 classification datasets with number of samples $\leq$ 10992 and number of features $\leq$ 100, and the 106 regression datasets with number of samples $\leq$ 8192 and number of features $\leq$ 100. Figure~\ref{fig:ds} presents a summary of dataset characteristics. Note that classification problems are both binary and multinomial.

\paragraph{Algorithms.}
We investigated 26 ML algorithms---13 classifiers and 13 regressors---using the following software packages: 
scikit-learn \cite{scikit-learn}, 
xgboost \cite{Chen:2016}, and 
lightgbm \cite{ke2017lightgbm}.
The algorithms are listed in Table~\ref{tab:hp}, along with the hyperparameter ranges or sets used in the hyperparameter search (described below).

\begin{table}
    \caption{Value ranges or sets used by Optuna for hyperparameter tuning. For ease of reference we use the function names of the respective software packages: scikit-learn, xgboost, and lightgbm. Values sampled from a range in the log domain are marked as `log', otherwise sampling is linear (uniform).}
    \label{tab:hp}
    \tiny
    \centering

    {\footnotesize\bf Classification}\\[5pt]
    
    \begin{tabular}{r|c|l}
    \hline
    \textbf{Algorithm} & \textbf{Hyperparameter} & \textbf{Values} \\ \hline
    \multirow{2}*{AdaBoostClassifier} & n\_estimators & [10, 1000]  (log) \\
     & learning\_rate & [0.1, 10]  (log) \\
    \hline
    \multirow{3}*{DecisionTreeClassifier} & max\_depth & [2, 10]  \\
     & min\_impurity\_decrease & [0.0, 0.5]  \\
     & criterion & \{gini, entropy\} \\
    \hline
    \multirow{3}*{GradientBoostingClassifier} & n\_estimators & [10, 1000]  (log) \\
     & learning\_rate & [0.01, 0.3]  \\
     & subsample & [0.1, 1]  \\
    \hline
    \multirow{3}*{KNeighborsClassifier} & weights & \{uniform, distance\} \\
     & algorithm & \{auto, ball\_tree, kd\_tree, brute\} \\
     & n\_neighbors & [2, 20]  \\
    \hline
    \multirow{3}*{LGBMClassifier} & n\_estimators & [10, 1000]  (log) \\
     & learning\_rate & [0.01, 0.2]  \\
     & bagging\_fraction & [0.5, 0.95]  \\
    \hline
    \multirow{3}*{LinearSVC} & max\_iter & [10, 10000]  (log) \\
     & tol & [1e-05, 0.1]  (log) \\
     & C & [0.01, 10]  (log) \\
    \hline
    \multirow{2}*{LogisticRegression} & penalty & \{l1, l2\} \\
     & solver & \{liblinear, saga\} \\
    \hline
    \multirow{2}*{MultinomialNB} & alpha & [0.01, 10]  (log) \\
     & fit\_prior & \{True, False\} \\
    \hline
    \multirow{3}*{PassiveAggressiveClassifier} & C & [0.01, 10]  (log) \\
     & fit\_intercept & \{True, False\} \\
     & max\_iter & [10, 1000]  (log) \\
    \hline
    \multirow{3}*{RandomForestClassifier} & n\_estimators & [10, 1000]  (log) \\
     & min\_weight\_fraction\_leaf & [0.0, 0.5]  \\
     & max\_features & \{auto, sqrt, log2\} \\
    \hline
    \multirow{2}*{RidgeClassifier} & solver & \{auto, svd, cholesky, lsqr, sparse\_cg, sag, saga\} \\
     & alpha & [0.001, 10]  (log) \\
    \hline
    \multirow{2}*{SGDClassifier} & penalty & \{l2, l1, elasticnet\} \\
     & alpha & [1e-05, 1]  (log) \\
    \hline
    \multirow{3}*{XGBClassifier} & n\_estimators & [10, 1000]  (log) \\
     & learning\_rate & [0.01, 0.2]  \\
     & gamma & [0.0, 0.4]  \\
    \hline
    \end{tabular}
    
    \normalsize
\end{table}

\begin{table}\ContinuedFloat
    \tiny
    \centering
    
    {\footnotesize\bf Regression}\\[5pt]
    
    \begin{tabular}{r|c|l}
    \hline
    \textbf{Algorithm} & \textbf{Hyperparameter} & \textbf{Values} \\ \hline
    \multirow{2}*{AdaBoostRegressor} & n\_estimators & [10, 1000]  (log) \\
     & learning\_rate & [0.1, 10]  (log) \\
    \hline
    \multirow{4}*{BayesianRidge} & n\_iter & [10, 1000]  (log) \\
     & alpha\_1 & [1e-07, 1e-05]  (log) \\
     & lambda\_1 & [1e-07, 1e-05]  (log) \\
     & tol & [1e-05, 0.1]  (log) \\
    \hline
    \multirow{3}*{DecisionTreeRegressor} & max\_depth & [2, 10]  \\
     & min\_impurity\_decrease & [0.0, 0.5]  \\
     & criterion & \{squared\_error, friedman\_mse, absolute\_error\} \\
    \hline
    \multirow{3}*{GradientBoostingRegressor} & n\_estimators & [10, 1000]  (log) \\
     & learning\_rate & [0.01, 0.3]  \\
     & subsample & [0.1, 1]  \\
    \hline
    \multirow{3}*{KNeighborsRegressor} & weights & \{uniform, distance\} \\
     & algorithm & \{auto, ball\_tree, kd\_tree, brute\} \\
     & n\_neighbors & [2, 20]  \\
    \hline
    \multirow{3}*{KernelRidge} & kernel & \{linear, poly, rbf, sigmoid\} \\
     & alpha & [0.1, 10]  (log) \\
     & gamma & [0.1, 10]  (log) \\
    \hline
    \multirow{3}*{LGBMRegressor} & lambda\_l1 & [1e-08, 10.0]  (log) \\
     & lambda\_l2 & [1e-08, 10.0]  (log) \\
     & num\_leaves & [2, 256]  \\
    \hline
    \multirow{2}*{LinearRegression} & fit\_intercept & \{True, False\} \\
     & normalize & \{True, False\} \\
    \hline
    \multirow{3}*{LinearSVR} & loss & \{epsilon\_insensitive, squared\_epsilon\_insensitive\} \\
     & tol & [1e-05, 0.1]  (log) \\
     & C & [0.01, 10]  (log) \\
    \hline
    \multirow{3}*{PassiveAggressiveRegressor} & C & [0.01, 10]  (log) \\
     & fit\_intercept & \{True, False\} \\
     & max\_iter & [10, 1000]  (log) \\
    \hline
    \multirow{3}*{RandomForestRegressor} & n\_estimators & [10, 1000]  (log) \\
     & min\_weight\_fraction\_leaf & [0.0, 0.5]  \\
     & max\_features & \{auto, sqrt, log2\} \\
    \hline
    \multirow{2}*{SGDRegressor} & alpha & [1e-05, 1]  (log) \\
     & penalty & \{l2, l1, elasticnet\} \\
    \hline
    \multirow{3}*{XGBRegressor} & n\_estimators & [10, 1000]  (log) \\
     & learning\_rate & [0.01, 0.2]  \\
     & gamma & [0.0, 0.4]  \\
    \hline
    \end{tabular}
    
    \normalsize
\end{table}

\paragraph{Metrics.}
We used three separate metrics for classification problems: 
\begin{enumerate}
    \item Accuracy: fraction of correct predictions; $\in [0,1]$.
    
    \item Balanced accuracy: an accuracy score that takes into account class imbalances, essentially the accuracy score with class-balanced sample weights \cite{sklearn-website}; $\in [0,1]$.
    
    \item F1 score: a harmonic mean of precision and recall; in the multi-class case, this is the average of the F1 score per class with weighting; $\in [0,1]$.
\end{enumerate}

We used three separate metrics for regression problems: 
\begin{enumerate}
    \item R2 score: $R^2$ (coefficient of determination) regression score function; $\in [-\infty,1]$.
    
    \item Adjusted R2 score: a modified version of the R2 score that adjusts for the number of predictors in a regression model. It is defined as $1 - (1 - r2) * (n - 1) / (n - p - 1)$, with $r2$ being the R2 score, $n$ being the number of samples, and $p$ being the number of features; $\in [-\infty,1]$.
  
    \item Complement RMSE: Complement of root mean squared error (RMSE), defined as $1- \mathit{RMSE}$; $\in [-\infty,1]$. This has the same range as the previous two metrics.
\end{enumerate}

\paragraph{Hyperparameter tuning.}
For hyperparameter tuning we used Optuna, a state-of-the-art automatic hyperparameter optimization software framework \cite{akiba2019optuna}. Optuna offers a define-by-run-style user API where one can dynamically construct the search space, and an efficient sampling algorithm and pruning algorithm. Moreover, our experience has shown it to be fairly easy to set up. Optuna formulates the hyperparameter optimization problem as a process of minimizing or maximizing an objective function that takes a set of hyperparameters as an input and returns its (validation) score. We used the default Tree-structured Parzen Estimator (TPE) Bayesian sampling algorithm. Optuna also provides pruning: automatic early stopping of unpromising trials \cite{akiba2019optuna}.

\paragraph{Overall flow.}
Algorithm~\ref{alg:setup} presents the top-level flow of the experimental setup.
For each combination of algorithm and dataset we perform 30 replicate runs. Each replicate separately assesses model performance over the respective three classification or regression metrics.
A replicate begins by splitting the dataset into training and test sets, and scaling them.
Then, per each metric: 
\begin{enumerate}
    \item Run Optuna over the training set for 50 trials to tune the model's hyperparameters, and retain the best model; compute the best model's test-set metric score.
    
    \item Evaluate 50 models over the training with default parameters and retain the best model; compute the best model's test-set metric score.\footnote{Strictly speaking, a few algorithms---Decision Tree, KNN, Bayesian---are essentially deterministic. For consistency we still performed the 50 default-hyperparameter trials. Further, our examining the respective implementations revealed possible randomness, e.g., for Decision Tree, when \textit{max\_features $<$ n\_features}, the algorithm will select \textit{max\_features} at random; though the default is \textit{max\_features $=$ n\_features} we still took no chances of there being some hidden randomness deep within the code.} 
\end{enumerate}
A model's evaluation is done through 5-fold cross-validation.
At the end of each replicate, compute the test-set percent improvement of Optuna's best model over the default's best model.

\begin{algorithm*}
\scriptsize
\caption{Experimental setup (per algorithm and dataset).}\label{alg:setup}
\begin{algorithmic}[1]
\Statex
\Require
\Indent
\Statex \textit{algorithm} $\gets$ algorithm to run
\Statex \textit{dataset} $\gets$ dataset to be used
\Statex \textit{n\_replicates} $\gets$ 30 (number of replicates)
\Statex \textit{n\_trials} $\gets$ 50 (number of Optuna trials, also number of runs with default values)
\Statex \textit{time\_limit} $\gets$ 72 hours (for all replicates)
\EndIndent

\Ensure
\Indent
\Statex Final scores (over test sets)
\EndIndent

\Statex
\Statex \LineComment{{\texttt{{\sq}metric1{\sq}}, \texttt{{\sq}metric2{\sq}}, \texttt{{\sq}metric3{\sq}}} are, respectively:}
\Statex \LineComment{\hspace{10pt} $\cdot$ For classification: accuracy, balanced accuracy, F1}
\Statex \LineComment{\hspace{10pt} $\cdot$ For regression: R2, adjusted R2, complement RMSE}
\Statex \LineComment{\textit{eval\_score} is 5-fold cross-validation score}

\Statex
\State Load \textit{dataset}
\For{\textit{rep} $\gets$ 1 to \textit{\textit{n\_replicates}}}
        \State Randomly split \textit{dataset} into 70\% \textit{training\_set} and 30\% \textit{test\_set}
        \State Fit \texttt{MinMaxScaler} to \textit{training\_set} and apply fitted scaler to \textit{training\_set} and \textit{test\_set}
        \For{\textit{metric} in {\texttt{{\sq}metric1{\sq}}, \texttt{{\sq}metric2{\sq}}, \texttt{{\sq}metric3{\sq}}}}
            \State Run Optuna with \textit{algorithm} for \textit{n\_trials} trials over \textit{training\_set} and obtain \textit{best\_model} \Comment{use \textit{eval\_score} for single-trial evaluation}
            \State Train \textit{best\_model} over \textit{training\_set}
            \State Compute \textit{metric} for \textit{best\_model} over \textit{test\_set}
            \For{\textit{i} in \textit{n\_trials}}
                \State Initialize a \textit{model} with default hyperparameters
                \State Evaluate \textit{model} over \textit{training\_set} using \textit{eval\_score}
                \If{\textit{eval\_score} is best obtained so far}
                \State Save \textit{model} as \textit{best\_model}
            \EndIf
        \EndFor
        \State Train \textit{best\_model} over \textit{training\_set}
        \State Compute \textit{metric} for \textit{best\_model} over \textit{test\_set}
    \EndFor
    \State \textit{imp1}, \textit{imp2}, \textit{imp3} = percent improvement Optuna over default for {\texttt{{\sq}metric1{\sq}}, \texttt{{\sq}metric2{\sq}}, \texttt{{\sq}metric3{\sq}}} \Comment{Compute and record replicate scores}
    \If{runtime $>$ \textit{time\_limit}}
        \State break
    \EndIf
\EndFor
\end{algorithmic}
\normalsize
\end{algorithm*} 

\section{Results and Discussion}
\label{sec:results}

A total of \reps replicates were performed, each comprising 300 algorithm runs (3 metrics $\times$ 50 Optuna trials, 3 metrics $\times$ 50 default trials), the final tally thus being \runs algorithm runs.\footnote{Note that for each run we called the \texttt{fit} method of the respective algorithm 5 times during 5-fold cross validation, i.e., the learning algorithm was executed 5 times.} 
%Due to minor post-submission code optimization (not bugs) and [reviewer..] we actually ran the entire experiment twice---witnessing identical qualitative results, with minor qualitative differences}
Table~\ref{tab:results} presents our results.

\begin{table}
    \caption{Compendium of final results over 26 ML algorithms, 250 datasets, \reps replicates, and \runs algorithm runs.
    A table row presents results of a single ML algorithm, showing a summary of all replicates and datasets.
    A table cell summarizes the results of an algorithm-metric pair.
    Cell values show \textit{median} and \textit{mean(std)}, where
    \textit{median}: median over all replicates and datasets of Optuna's percent improvement over default;
    \textit{mean(std)}: mean (with standard deviation) over all replicates and datasets of Optuna's percent improvement over default.
    Also shown is total number of replicates for which these statistics were computed.
    Acc: accuracy score; Bal: balanced accuracy score; F1: F1 score; R2: R2 score; Adj R2: adjusted R2 score; C-RMSE: complement RMSE; Reps: total number of replicates.
    NB: Number of replicates may be smaller than the maximal possible value: $144 \times 30 = 4320$ for classification datasets, and $106 \times 30 = 3180$ for regression datasets. This is due to edge cases that cause a single replicate to terminate with an error, the vicissitudes of life on the cluster, and in small part to long runtimes evoking the 72-hour timeout (this happened with GradientBoostingClassifier for 14 datasets and with XGBClassifier for 8 datasets).}
    \label{tab:results}
    % \ssmall
    \tiny
    \vspace{5pt}
    \centering

    {\small\bf Classification} \\[5pt]
    
    \begin{tabular}{r|cc|cc|cc|c}
        \hline
        \multirow{2}*{\textbf{Algorithm}} & \multicolumn{2}{c|}{\textbf{Acc}} & \multicolumn{2}{c|}{\textbf{Bal}} & \multicolumn{2}{c|}{\textbf{F1}} & \textbf{Reps} \\ 
        & \textit{median} & \textit{mean(std)}  & \textit{median} & \textit{mean(std)}  & \textit{median} & \textit{mean(std)} & \\ \hline
         AdaBoostClassifier & 1.9 & 20.9 (65.3) & 2.2 & 21.5 (57.4) & 1.9 & 39.3 (150.1) & 4320 \\ \hline
         DecisionTreeClassifier & 0.0 & 115.6 (2.3e+03) & 0.0 & 96.0 (2.2e+03) & 0.0 & 55.7 (1.3e+03) & 4220 \\ \hline
         GradientBoostingClassifier & 0.5 & 45.1 (1.4e+03) & 0.6 & 48.9 (1.4e+03) & 0.6 & 42.0 (1.1e+03) & 4096 \\ \hline
         KNeighborsClassifier & 0.8 & 3.8 (13.8) & 1.8 & 5.9 (16.7) & 1.5 & 5.1 (17.7) & 4254 \\ \hline
         LGBMClassifier & 0.0 & 1.2 (11.5) & 0.0 & 1.0 (11.9) & 0.0 & 0.9 (12.1) & 4287 \\ \hline
         LinearSVC & 0.0 & 1.0 (8.3) & 0.0 & 1.9 (8.5) & 0.0 & 1.7 (8.2) & 4299 \\ \hline
         LogisticRegression & 0.0 & 1.5 (8.2) & 0.0 & 3.4 (12.5) & 0.0 & 3.4 (11.8) & 4307 \\ \hline
         MultinomialNB & 0.0 & 9.8 (58.9) & 8.5 & 27.5 (48.9) & 10.5 & 40.5 (128.6) & 4149 \\ \hline
         PassiveAggressiveClassifier & 1.9 & 7.8 (24.0) & 1.8 & 5.9 (18.6) & 3.0 & 10.8 (28.8) & 4301 \\ \hline
         RandomForestClassifier & 0.0 & 153.6 (2.3e+03) & 0.0 & 218.8 (3.0e+03) & 0.0 & 134.1 (2.0e+03) & 4320 \\ \hline
         RidgeClassifier & 0.0 & 1.0 (6.8) & 0.0 & 1.4 (7.3) & 0.0 & 1.9 (7.9) & 4273 \\ \hline
         SGDClassifier & 1.2 & 5.0 (20.6) & 1.6 & 5.2 (16.7) & 2.0 & 8.6 (26.7) & 4212 \\ \hline
         XGBClassifier & 0.0 & 13.5 (643.1) & 0.0 & 11.4 (431.2) & 0.0 & 10.1 (467.6) & 4111 \\ \hline
    \end{tabular} \\[15pt]

    {\small\bf Regression} \\[5pt]
         
    \begin{tabular}{r|cc|cc|cc|c}
        \hline
        \multirow{2}*{\textbf{Algorithm}} & \multicolumn{2}{c|}{\textbf{R2}} & \multicolumn{2}{c|}{\textbf{Adj}} & \multicolumn{2}{c|}{\textbf{C-RMSE}} & \textbf{Reps} \\ 
        & \textit{median} & \textit{mean(std)}  & \textit{median} & \textit{mean(std)}  & \textit{median} & \textit{mean(std)} & \\ \hline
         AdaBoostRegressor & 2.0 & 3.6 (33.5) & 2.1 & -741.3 (9.5e+03) & 3.8 & 5.1 (20.9) & 3179 \\ \hline
         BayesianRidge & 0.0 & 6.8e+03 (3.7e+05) & -0.0 & -3.3 (55.3) & 0.0 & 1.0 (9.8) & 3117 \\ \hline
         DecisionTreeRegressor & 3.8 & 61.5 (788.0) & 4.0 & 49.1 (841.5) & 7.0 & 63.4 (1.3e+03) & 3150 \\ \hline
         GradientBoostingRegressor & 1.6 & 17.3 (430.9) & 1.7 & -6.6e+05 (2.4e+07) & 4.1 & 2.3 (126.4) & 3180 \\ \hline
         KNeighborsRegressor & 3.5 & 77.8 (627.4) & 3.5 & 18.8 (471.6) & 4.5 & 203.5 (5.3e+03) & 3160 \\ \hline
         KernelRidge & 69.5 & -9.3e+05 (5.0e+07) & 65.9 & 3.6e+03 (1.7e+05) & 49.5 & 1.7e+03 (8.1e+04) & 3053 \\ \hline
         LGBMRegressor & 0.0 & 0.0 (25.6) & 0.0 & -1.2 (34.4) & 0.0 & 0.4 (2.1) & 3179 \\ \hline
         LinearRegression & 0.0 & 2.3 (70.4) & 0.0 & -35.1 (469.4) & 0.0 & -1.7 (62.8) & 3170 \\ \hline
         LinearSVR & 25.1 & 86.4 (2.7e+03) & 24.3 & 173.5 (2.8e+03) & 23.9 & 159.7 (2.2e+03) & 3161 \\ \hline
         PassiveAggressiveRegressor & 71.6 & 180.7 (1.7e+03) & 58.5 & -304.3 (4.1e+03) & 62.0 & 331.9 (5.5e+03) & 3167 \\ \hline
         RandomForestRegressor & -0.1 & 1.5 (44.2) & -0.2 & -1.2e+03 (4.6e+04) & -0.5 & -1.5 (13.2) & 3180 \\ \hline
         SGDRegressor & 0.0 & 2.6 (68.6) & 0.0 & -41.4 (2.0e+03) & 0.0 & 2.2 (39.8) & 3167 \\ \hline
         XGBRegressor & 0.9 & 20.0 (717.1) & 0.8 & -675.6 (7.4e+03) & 2.3 & 6.8 (164.6) & 3180 \\ \hline
    \end{tabular}
    
    \normalsize
\end{table}

Examining Table~\ref{tab:results} brings to the fore several interesting points. First, regressors are somewhat more susceptible to hyperparameter tuning, i.e., there is more to be gained by tuning vis-a-vis the default hyperparameters.
    
For most classifiers and---to a lesser extent---regressors, the median value shows little to be gained from tuning; yet the mean value along with the standard deviation suggests that for some algorithms there is a wide range in terms of tuning effectiveness. Indeed, examining the collected raw experimental results, we noted that at times there was a ``low-hanging fruit'' case: the default hyperparameters yielded very poor performance on some datasets, leaving room for considerable improvement through tuning.

It would seem useful to define a ``bottom-line'' measure---a summary score, as it were, which essentially summarizes an entire table row, i.e., an ML algorithm's sensitivity to hyperparameter tuning. We believe any such measure would be inherently arbitrary to some extent; that said, we nonetheless put forward the following definition of \textit{hp\_score}:
\begin{itemize}
    \item Consider (separately for classifiers and regressors) the 13 algorithms and 9 measures of Table~\ref{tab:results} as a dataset with 13 samples and the following 9 features: 
        \textit{metric1\_median}, 
        \textit{metric2\_median}, 
        \textit{metric3\_median},  
        \textit{metric1\_mean}, 
        \textit{metric2\_mean}, 
        \textit{metric3\_mean}, 
        \textit{metric1\_std}, 
        \textit{metric2\_std}, 
        \textit{metric3\_std}
        
    \item Apply scikit-learn's RobustScaler, which scales features using statistics that are robust to outliers: ``This Scaler removes the median and scales the data according to the quantile range (defaults to IQR: Interquartile Range). The IQR is the range between the 1st quartile (25th quantile) and the 3rd quartile (75th quantile). Centering and scaling happen independently on each feature...'' \cite{sklearn-website}
    
    \item The \textit{hp\_score} of an algorithm is then simply the mean of its 9 scaled features.
\end{itemize}
This \textit{hp\_score} is unbounded because improvements or impairments can be arbitrarily high or low. A higher value means that the algorithm is expected to gain more from hyperparameter tuning, while a lower value means that the algorithm is expected to gain less from hyperparameter tuning  (on average).

Table~\ref{tab:hpscore} presents the \textit{hp\_score}s of all 26 algorithms, sorted from highest to lowest per algorithm category (classifier or regressor). While simple and immanently imperfect, \textit{hp\_score} nonetheless seems to summarize fairly well the trends observable in Table~\ref{tab:results}. 

\begin{table}
    \caption{The \textit{hp\_score} of each ML algorithm, computed from the values in Table~\ref{tab:results}. A higher value means that the algorithm is expected to gain more from hyperparameter tuning, while a lower value means that the algorithm is expected to gain less from hyperparameter tuning.}
    \label{tab:hpscore}
    \footnotesize
    \vspace{5pt}
    \centering
    
    \begin{tabular}{c|c}
    {\small\bf Classification} & {\small\bf Regression} \\ \hline
    \begin{tabular}{rl}
         RandomForestClassifier & 3.89 \\
         DecisionTreeClassifier & 2.43 \\
         GradientBoostingClassifier & 1.52 \\
         MultinomialNB & 1.36 \\
         AdaBoostClassifier & 0.79 \\
         PassiveAggressiveClassifier & 0.56 \\
         XGBClassifier & 0.38 \\
         SGDClassifier & 0.35 \\
         KNeighborsClassifier & 0.27 \\
         LogisticRegression & -0.08 \\
         LinearSVC & -0.09 \\
         RidgeClassifier & -0.10 \\
         LGBMClassifier & -0.10 \\
    \end{tabular} 
    &
    \begin{tabular}{rl}
         KernelRidge & 2110.75 \\
         GradientBoostingRegressor & 183.97 \\
         BayesianRidge & 35.19 \\
         PassiveAggressiveRegressor & 5.34 \\
         LinearSVR & 2.13 \\
         KNeighborsRegressor & 0.59 \\
         DecisionTreeRegressor & 0.35 \\
         RandomForestRegressor & 0.09 \\
         AdaBoostRegressor & -0.07 \\
         XGBRegressor & -0.10 \\
         SGDRegressor & -0.23 \\
         LinearRegression & -0.25 \\
         LGBMRegressor & -0.26 \\
    \end{tabular}
    
    \end{tabular}
    
\normalsize
\end{table}

The main insight from Tables~\ref{tab:results} and~\ref{tab:hpscore} is:
For most ML algorithms, we should not expect huge gains from hyperparameter tuning \textit{on average}, however, there may be some datasets for which default hyperparameters perform poorly, this latter being truer for some algorithms than others. In particular, those algorithms at the bottom of the lists in Table~\ref{tab:hpscore} would likely not benefit greatly from a significant investment in hyperparameter tuning. Some algorithms are robust to hyperparameter selection, others somewhat less.

Table~\ref{tab:hpscore} can be used in practice by an ML practitioner: 1) to decide how much to invest in hyperparameter tuning of a particular algorithm, and 2) to select algorithms that require less tuning. Hopefully, this will save time and energy \cite{GARCIAMARTIN201975}.
For algorithms at the top of the lists, we may inquire as to whether particular hyperparameters are the root cause of their hyperparameter sensitivity; further, we may seek out better defaults. For example, \cite{stuke2021efficient} recently focused on hyperparameter tuning for KernelRidge, which is at the top of the regressor list in Table~\ref{tab:hpscore}.
\cite{Probst2019} discussed the tunability of a specific hyperparameter, though they noted the problem of hyperparameter dependency.

\section{Concluding Remarks}
\label{sec:disc}
We performed a large-scale experiment of hyperparameter-tuning effectiveness, across multiple ML algorithms and datasets. More algorithms can be added in the future, as well as more datasets and more hyperparameters. Further, one can consider tweaking specific components of the setup (e.g., the metrics and the scaler of Algorithm~\ref{alg:setup}). The code is available at \url{https://github.com/moshesipper}). 

Given the findings herein, it seems that more often than not hyperparameter tuning will not provide huge gains over the default hyperparameters of the respective software packages examined (in addition to our findings it also appears that most defaults were judiciously selected). A modicum of tuning would seem to be advisable, though other factors will likely play a stronger role in final model performance, including, to name a few: quality of raw data,  solidity  of data preprocessing, and choice of ML algorithm (curiously, this latter can be considered a tunable hyperparameter \cite{SipperAddGBoost2022}).

\section*{Acknowledgements}
I thank Raz Lapid for helpful comments.

\small
\bibliography{refs}
\bibliographystyle{plain}
\normalsize

\end{document}